\documentclass[lettersize,journal]{IEEEtran}
\usepackage{graphicx} %use graph format
\usepackage[percent]{overpic}
\usepackage{epstopdf}
\usepackage{amsmath,amsfonts}
\usepackage{algorithmic}
\usepackage{amsthm}
\usepackage{algorithm}
\usepackage{array}
\usepackage[dvipsnames, svgnames, x11names,table,xcdraw]{xcolor}
\usepackage{tikz}
\usepackage{multirow}
\usepackage{comment}
\usepackage{textcomp}
\usepackage{stfloats}
\usepackage{url}
\usepackage{verbatim}
\usepackage{amssymb}

\usepackage{makecell}
\usepackage{booktabs}
\usepackage[normalem]{ulem}
\useunder{\uline}{\ul}{}
\usepackage{bm}
\usepackage{adjustbox}
\usepackage{tabularx}
\usepackage[caption=false,font=normalsize,labelfont=sf,textfont=sf]{subfig}
\usepackage{float}
\usepackage{needspace}
\usepackage[hidelinks]{hyperref}
\DeclareRobustCommand{\authoremail}[1]{{\urlstyle{same}\nolinkurl{#1}}}
\definecolor{bg01}{RGB}{240,240,255} % 淡紫蓝
\definecolor{bg02}{RGB}{255,240,240} % 淡粉
\definecolor{bg03}{RGB}{255,255,230} 

\definecolor{headerblue}{RGB}{220, 235, 250}
\definecolor{rowours}{RGB}{235, 248, 240}
\definecolor{bestcell}{RGB}{200, 230, 200}
\usepackage{cite}

\begin{document}
\setlength{\belowdisplayskip}{4pt}
\title{Towards Seconds-level Video Representation: Quadratically Gated Feature Grid with Adaptive, Prediction-free Sampling}

\author{Tie-Yun Li$^{1}$, Hao-Ran Wang$^{2,*}$, Chao Wang$^{2}$, Zhiyong Yin$^{3}$, Yiming Xu$^{3}$, Zhenwei Cai$^{3}$,\\ Li Li$^{4}$, Rongzhi Lin$^{1}$, Feng-Lei Fan$^{2}$

\thanks{$^{*}$Corresponding author: Hao-Ran Wang (\authoremail{hwang965-c@my.cityu.edu.hk})}
\thanks{$^{1}$Tie-Yun Li (\authoremail{tieyun.li@stu.hit.edu.cn}) and Rongzhi Lin (\authoremail{linrongzhi@hit.edu.cn}) are with the School of Mathematics, Harbin Institute of Technology, Harbin 150001, China.}
\thanks{$^{2}$Hao-Ran Wang, Chao Wang (\authoremail{wangchao.hit@gmail.com}), and Feng-Lei Fan (\authoremail{fenglfan@cityu.edu.hk}) are with Frontier of Artificial Networks (FAN) Lab, Department of Data Science,
City University of Hong Kong, Hong Kong, China SAR.}
\thanks{$^{3}$Zhiyong Yin (\authoremail{ryan_yinzy@163.com}), Yiming Xu (\authoremail{ycxuyiming@gmail.com}), and Zhenwei Cai (\authoremail{741281990@qq.com}) are with China Construction Third Engineering Bureau (Shenzhen) Co., Ltd., Shenzhen, Guangdong 518110, China.}
\thanks{$^{4}$Li Li (\authoremail{lil1@ustc.edu.cn}) is with the MoE Key Laboratory of Brain-Inspired Intelligent Perception and Cognition, University of Science and Technology of China, Hefei 230027, China.}
}

\maketitle

\begin{abstract}

Implicit neural video representations are notoriously slow to build, which does not cross the basic practical threshold for most industrial applications. Here, we present Q-GAS, a coordinate-queryable video representation that couples multi-resolution explicit 3D feature grids with a compact SwiGLU-style quadratic-gating (QG) decoder and an error-aware, prediction-free sampler. The explicit grids avoid hash-collision coefficient sharing at the cost of modestly higher storage yet greatly boost the expressive capacity; the QG decoder introduces low-rank local second-order interactions to improve the nonlinear feature response. During fitting, a global error map periodically refreshes a high-error pool, and each batch mixes pool samples with uniform samples, which is free from the auxiliary model to predict difficult samples, thereby further escalating the efficiency. On UVG Beauty/Bosphorus/Jockey, Q-GAS fits and quantizes under 50s on one RTX PRO 6000, decodes at over 370 FPS with mixed precision and about 146 FPS in FP32. In our PyTorch implementation, Q-GAS trains and quantizes 5.5–7.2$\times$ faster than the evaluated Instant-NGP baseline and 19–28$\times$ faster than RNeRV. Under a 90\,s budget, it improves PSNR/MS-SSIM/LPIPS over the evaluated NeRV-family pipelines at comparable BPP over the measured overlap. We have shared our code at \url{https://github.com/MillenRosen/Q-GAS} for readers' free use.
\end{abstract}


\section{Introduction}

\IEEEPARstart{V}{ideo} contents account for the majority of the Internet traffic, necessitating efficient representation methods \cite{cisco2019vni, bross2021overview, lu2019dvc}. Implicit neural representations (INRs) have recently shown promising results in video representation \cite{nerv, enerv, hnerv, hinerv2023}, which model a 3D video signal as a continuous function $f_{\theta}: (x,y,t)\to(r,g,b)$, thereby enabling queries at arbitrary spatial and temporal coordinates \cite{nerf, siren, coin}. However, a coordinate-based representation through a layered network is computationally expensive for videos. This large cost not only makes the training slow, since the parameters of INRs must be fitted to the entire video \cite{gwilliam2025design, instantngp}, but also delays queries. As a consequence, the wide adoption of INRs in applications that require a quality video representation to be obtained quickly is hindered, \textit{e.g.}, video compression. To alleviate this problem, there emerge two lineage-related but methodologically distinct schools of INRs: the Neural Representations for Videos (NeRVs) and hybrid representations. 

Among the NeRV methods, NeRV \cite{nerv}, ENeRV \cite{enerv}, and HNeRV \cite{hnerv} improve the throughput by generating full frames from temporal or content embeddings. Subsequent architectures incorporate learned feature grids, hierarchical features, and motion models \cite{lee2023ffnerv,hinerv2023,dsnerv2024}. Recently, RNeRV \cite{gwilliam2025design} makes a further step by modularizing the existing NeRV models and recommending the optimal portfolio of different modules given different training time. These NeRV variants speed up video representation at both the decoding and encoding phases because they shift from per-pixel implicit queries to frame-level generation: a single forward pass of a convolutional decoder maps a frame index to the entire RGB image. Yet encoding still requires substantial optimization for each video, \textit{e.g.}, it still takes over ten minutes to fit an 8-bit color depth video with around 1.7GB storage. In short, NeRV solves the slow query problem, but is handicapped in fitting.

Hybrid representations support coordinate queries with a compact decoder. They store part of the signal representation in learnable grids or factorized structures and interpolate local features from them \cite{instantngp,tensorf,fridovich2022plenoxels,kplanes2023,hexplane2023}. Using these features with a neural decoder requires careful allocation of storage and computational resources \cite{instantngp,tensorf}. Finer grids provide more independent coefficients to represent local detail but increase storage, whereas a larger decoder increases the computational cost of each query. The central challenge of hybrid representations is to recover sufficient details from compact features with an emphasis on poorly-reconstructed regions during training \cite{martel2021acorn,katharopoulos2018not}. Encoding and sampling must therefore be considered together when designing a representation for rapid fitting. Compared to NeRV, the design of hybrid representations reduces the per-video fitting burden while retaining coordinate-query support \cite{nerv,nvp2022}. Nevertheless, fitting an 8-bit video of around 1.7 GB still takes about six minutes. 

\begin{figure*}[!htbp]
\centering
\includegraphics[width=\textwidth]{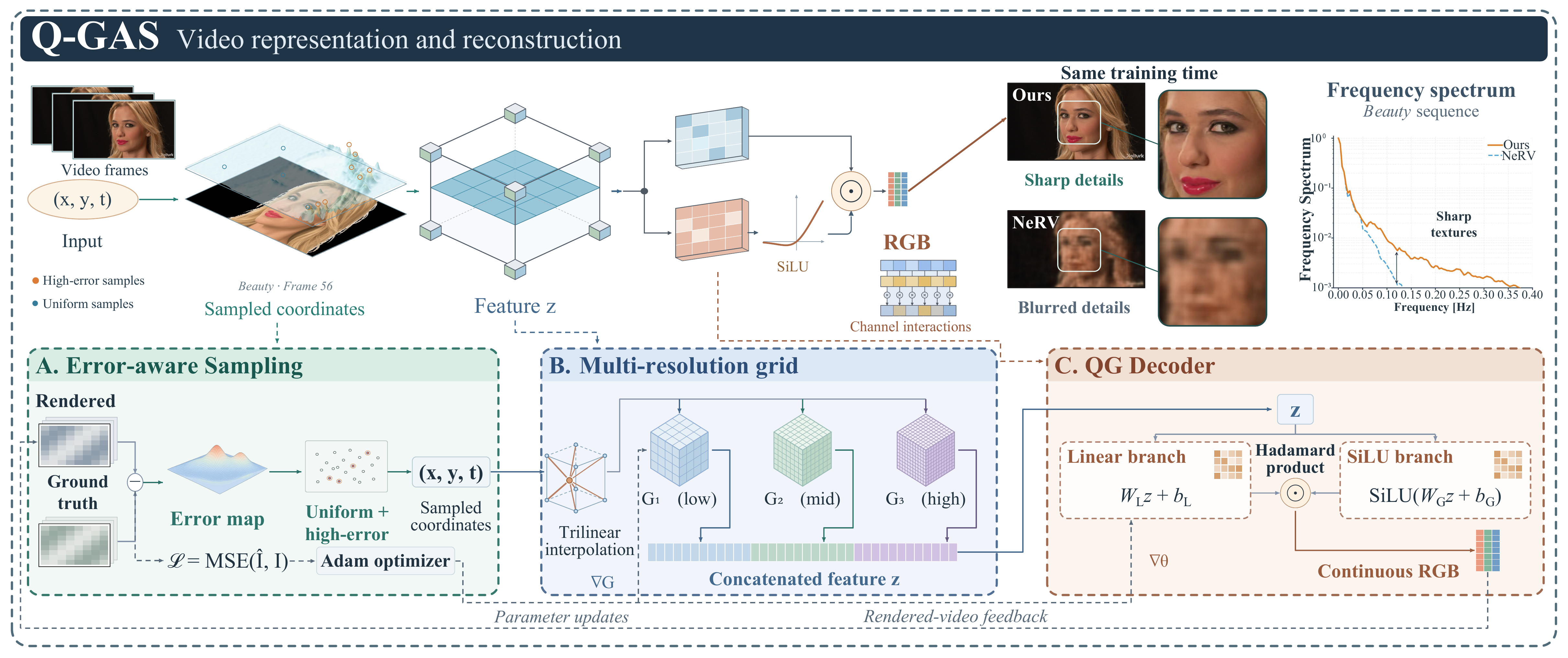}
\caption{An overview of Q-GAS.  \textbf{Multi-Resolution 3D Feature Grids:} At each coordinate $(x,y,t)$, trilinear interpolation retrieves features. The schematic illustrates the multiscale construction.  \textbf{QG Decoder:} A linear branch and a SiLU-gated branch quadratically interact through a Hadamard product, which serves as a compact RGB decoder. \textbf{Error-aware Sampling:} Reconstruction errors constitute the global error map during training, and Adam optimizes the representation with uniform  coordinates and coordinates from the current high-error pool.}
\label{fig:teaser}
\vspace{-0.2cm}

\end{figure*}

Although INRs enjoy unprecedented flexibility compared to other video codecs, if too slowly encoded, they cannot participate in any workflow where video is ingested, processed, and delivered continuously. Currently, industrial video pipelines operate on latency budgets far tighter than the encoding times achieved by existing INR methods, \textit{e.g.}, cloud-gaming motion-to-photon latency targets are around 100 ms, and JPEG XS is standardized for sub-frame, near-zero-latency broadcast and pro-AV workflows \cite{descampe2021jpegxs}. \textit{Can we accelerate INRs for videos to the level of seconds?}  Though fitting in tens of seconds remains too slow for real-time applications, it crosses the basic practical threshold for latency-tolerant video ingestion, on-demand offline analysis, and preparation for interactive video editing \cite{xu2019vstore,hsieh2018focus,huang2023inve}. To address this question, as Figure~\ref{fig:teaser} shows, we propose \underline{\textbf{Q}}uadratically-gated \underline{\textbf{G}}rid Representation with \underline{\textbf{A}}daptive, Prediction-free \underline{\textbf{S}}ampling (Q-GAS), a continuous video representation built on quadratically gated explicit multi-resolution spatiotemporal feature grids and prediction-free sampling. Q-GAS can accomplish the normative continuous video representation in seconds while retaining the representation quality through two levels of innovations:

At the representation level, Q-GAS couples explicit multi-resolution 3D feature grids with a compact quadratically-gated (QG) decoder. Specifically, i) 3D feature grids assign an independent learnable feature vector to each lattice vertex at each resolution. Trilinear interpolation retrieves these local features without introducing collision-induced feature sharing among distinct grid vertices like hash-grid encoding. Our theoretical analyses indicate that the proposed 3D feature grid preserves the better expressive capacity of the spatial grid for local multiscale feature encoding than Hash-based coding. ii) The QG decoder maps concatenated grid features to RGB values through a compact quadratically-gated network, which uses a low-rank quadratic architecture in which an affine content branch and a SiLU-activated gating branch interact through a Hadamard product. These multiplicative interactions allow the decoder to modulate feature responses and capture dependencies across feature channels to recover high-frequency details. Moreover, due to the low-rank nature of the quadratic gate, the decoder improves reconstruction quality without substantially increasing parameter count.

At the training level, Q-GAS uses an error-aware yet prediction-free coordinate sampling to further accelerate fitting. During fitting, one maintains a global spatiotemporal error pool using reconstruction errors between the representation and the ground truth. It combines coordinates sampled uniformly across the input with coordinates drawn from a periodically refreshed high-error pool. This directs additional updates toward regions with large errors in the premise of maintaining global coverage. Because the feedback comes directly from reconstruction errors, the coordinate sampler adapts to the evolving representation without requiring an auxiliary prediction network or explicit motion estimation.

The representation and training components seamlessly complement each other to support rapid fitting. Prediction-free sampler effectively determines \textit{where} to allocate optimization efforts, while the QG decoder determines \textit{how} to transform the sampled features into RGB values. Multiplicative interactions increase the decoder's expressive capacity and can recover fine detail with few parameters. \textbf{
Experiments on \textit{Beauty} and \textit{Bosphorus} show that Q-GAS achieves PSNR above 32 dB, completes training and quantization within 50\,s, and decodes at over 370 FPS using mixed precision.
} Specifically, compared with the evaluated Instant-NGP implementation, Q-GAS completes training and quantization more than 5$\times$ faster and decodes at more than 20$\times$ its throughput. Meanwhile, under a fixed training time and comparable BPP over the overlapping measured ranges, Q-GAS achieves higher PSNR, SSIM, and MS-SSIM than the evaluated NeRV-family pipelines, while maintaining real-time FP32 decoding. Our contributions are threefold:

\begin{itemize}
\item At the representation level, we introduce Q-GAS, a fast implicit neural video representation that couples multi-resolution 3D feature grids with a compact gated readout incorporating low-rank quadratic feature interactions.

\item At the training level, we introduce a dynamic error-aware yet prediction-free sampling mechanism that directs additional training samples toward video regions that remain poorly reconstructed as the representation is optimized. To the best of our knowledge, Q-GAS is the first INR for videos that uses this kind of sampling strategy.

\item Empirically, systematic experiments confirm that Q-GAS can represent a standard video in tens of seconds, which is the record-high speed in the field of INRs for videos.

\end{itemize}

\section{Related Work}

\subsection{Advances in INRs}
Coordinate-based INRs use a neural network to parameterize a signal as a continuous function from coordinates to signal values. SIREN \cite{siren} employs periodic activations to capture fine details and allow signal derivatives to be computed analytically, while Fourier feature mappings \cite{fourierfeatures} improve the ability of coordinate multilayer perceptrons (MLPs) to represent high-frequency functions. Coordinate-based INRs can also yield compact signal representations through compressing parameters. For example, COIN \cite{coin} stores an image entirely in quantized network parameters. However, this coordinate-based formulation comes at a cost: the signal is represented solely through network parameters, so each query requires a full network evaluation. This makes per-coordinate evaluation costly for high-resolution video, especially with deep decoders. To reduce this cost, we design Q-GAS, an efficient neural representation which retains coordinate queries while storing part of the representation in explicit feature grids.

An alternative to per-coordinate querying is to generate each frame directly from a temporal index, trading coordinate-level flexibility for a high rendering throughput. NeRV \cite{nerv} is a representative method along this direction: it passes a temporal index through convolutional blocks and PixelShuffle upsampling to produce a full frame. Later designs improve this frame-generation paradigm by enriching features, disentangling spatial and temporal context, or exploiting temporal redundancy. ENeRV \cite{enerv} disentangles spatial and temporal context, while HNeRV \cite{hnerv} introduces content-adaptive embeddings. FFNeRV \cite{lee2023ffnerv} models optical flow to exploit temporal redundancy, and HiNeRV \cite{hinerv2023} combines hierarchical features with convolutional decoding for video compression. DS-NeRV \cite{dsnerv2024} separates static and dynamic codes through cross-channel attention. RNeRV \cite{gwilliam2025design} combines components from the NeRV-family to improve the trade-off between quality, model size, and training time. It uses learned feature grids as an auxiliary feature source for a convolutional frame generator and produces each frame as a whole. In contrast, our model treats multi-resolution 3D feature grids as part of the representation itself: it queries them at individual spatiotemporal coordinates and decodes the locally interpolated features with a compact QG decoder. Compared with NeRV-family frame generation, this design retains coordinate-level query flexibility while keeping both fitting and per-query decoding costs low.

Our grid-decoder design is a type of hybrid representation which aims to alleviate the cost of full MLP evaluation at each coordinate by storing latent codes in explicit geometric structures. Instant-NGP \cite{instantngp} is a widely used hybrid representation method with multi-resolution hash tables and a small neural network to fit images and 3D fields rapidly. For volumetric and spatiotemporal modeling, methods such as K-Planes \cite{kplanes2023} and HexPlane \cite{hexplane2023} use orthogonal 2D planes, while DVGO \cite{sun2022dvgo} and TiNeuVox \cite{fang2022tineuvox} use explicit voxels to represent the underlying scene efficiently. Building on this hybrid formulation, Q-GAS stores features in multi-resolution 3D feature grids, queries them via the trilinear interpolation, and learns quadratic interactions between feature channels in the compact QG decoder. The main contribution of our method is to present a super-fast yet effective representation and training strategy, rather than a new interpolation kernel.

\subsection{Quadratic Networks and Parameter Efficiency}
Quadratic neurons incorporate second-order interactions into the aggregation function of a neuron, allowing individual neurons to model nonlinear relationships between input features. Fan et al. \cite{fan2025one} theoretically demonstrate that quadratic networks can achieve higher parameter efficiency than conventional neural networks under specific function approximation settings. In imaging and signal reconstruction, quadratic models have been applied to low-dose CT denoising \cite{fan2020quadratic}, intracranial hemorrhage localization \cite{ferdi2024quadratic}, and seismic noise attenuation \cite{wang2025quadratic}. Quadratic residual networks further demonstrate their potential for parameter-efficient function approximation in forward and inverse PDE problems \cite{bu2021quadratic}. Although quadratic models can express richer feature interactions, they incur additional computational overhead. QuadraNet-V2 \cite{xu2026quadranet} alleviates this problem by reusing pretrained linear components and learning quadratic adaptations.

Motivated by the success of quadratic networks in image processing, we introduce second-order feature interactions into INR-based video representations through Q-GAS. Recent work has also applied higher-order implicit representations to improve high-frequency signal modeling in imaging inverse problems \cite{chen2025hope}. In contrast, our work extends quadratic networks to implicit neural video representation, achieving high-fidelity reconstruction with minimal parameter overhead. Specifically, Q-GAS applies SwiGLU-style gating to features sampled from explicit multi-resolution 3D grids for coordinate-based video reconstruction. Its local quadratic response captures interactions between two projection directions using a number of parameters that scales linearly with the dimension of the concatenated grid features.

\subsection{Importance Sampling in Neural Optimization}
Uniform coordinate sampling provides broad coverage of a video sequence but may allocate training samples to regions that the model already reconstructs accurately. Allocating more updates to difficult examples is a well-established training strategy, as in Online Hard Example Mining (OHEM) for object detection \cite{shrivastava2016training}. In coordinate-based neural representations, some adaptive sampling methods employ additional networks or spatial partitioning mechanisms to guide sample allocation. For instance, Neural Radiance Fields (NeRF) \cite{nerf} uses predictions from a coarse network to guide the allocation of fine samples along each camera ray. The Adaptive Coordinate Network (ACORN) \cite{martel2021acorn} refines a multiscale partition of the coordinate domain based on local fitting errors, allocating more samples and model capacity to difficult regions. These methods illustrate the benefits of adaptive resource allocation, but the auxiliary networks or partitioning procedures introduce additional computational and implementation costs.

% For rapid video fitting, however, the benefit of such auxiliary components must be balanced against their additional computational cost. 
In comparison, Q-GAS adopts an error-aware coordinate sampling strategy for rapid video fitting without introducing an auxiliary neural network, which prioritizes difficult yet important coordinates while limiting the cost of maintaining sampling priorities. The sampler updates a global spatiotemporal error map using reconstruction errors during training and periodically refreshes a pool of high-error coordinates. Each training batch mixes uniformly sampled coordinates with samples drawn from this pool to maintain sequence coverage while emphasizing poorly reconstructed regions.

\section{Methodology}

Q-GAS represents a video using local features interpolated from multi-resolution 3D feature grids and a quadratically gated decoder, as illustrated in Figure~\ref{fig:teaser}. Let $\mathbf{V}\in[0,1]^{T\times H\times W\times3}$ denote a video containing $T$ RGB frames of spatial resolution $H\times W$, and let $\mathbf{x}=(x,y,t)^\top\in\Omega=[-1,1]^3$ denote a normalized spatiotemporal query coordinate. Each query accesses $L$ explicit 3D grids at different resolutions and retrieves local features by trilinear interpolation. The $l$-th grid produces a feature vector $\mathbf{f}_l(\mathbf{x})\in\mathbb{R}^{C}$. We concatenate \(L\) feature vectors into \(\mathbf{z}(\mathbf{x})\in\mathbb{R}^{LC}\) and predict RGB values using two QG layers \(g_1,g_2\) followed by a final affine projection:

\begin{equation}
\begin{aligned}
\mathbf{z}(\mathbf{x})&=\operatorname{Concat}\bigl(\mathbf{f}_1(\mathbf{x}),\ldots,\mathbf{f}_L(\mathbf{x})\bigr),\\
\widehat{\mathbf{V}}(\mathbf{x})&=\sigma\!\left(\mathbf{W}_{\mathrm{out}}g_2\!\left(g_1(\mathbf{z}(\mathbf{x}))\right)+\mathbf{b}_{\mathrm{out}}\right).
\end{aligned}
\label{eq:multires_feature}
\end{equation}
Here $\sigma$ is the logistic sigmoid, which bounds the output RGB values. To fit this representation to the observed video, we jointly optimize the grid coefficients and decoder parameters by minimizing the mean squared error over a batch $\mathcal{B}$ of observed pixel coordinates:

\begin{equation}
\mathcal{L} = \frac{1}{3|\mathcal{B}|}\sum\nolimits_{\mathbf{x}\in\mathcal{B}}\left\|\widehat{\mathbf{V}}(\mathbf{x})-\mathbf{V}(\mathbf{x})\right\|_2^2.
\label{eq:reconstruction_loss}
\end{equation}

The sampling policy determines the coordinates used in each training batch, while the reconstruction loss assigns equal weight to every sampled coordinate. Under the dynamic sampling policy, observed residuals update an error map that guides subsequent coordinate selection during fitting. After fitting, we quantize and package the grid coefficients and decoder parameters together with the configuration needed for reconstruction, such as resolution and quantization parameters. Decoding loads this stored representation and evaluates the grid--decoder mapping at the requested coordinates. The training error map and sampling pools are not needed for reconstruction. We now describe the multi-resolution grid representation, the QG decoder, and the error-aware coordinate sampling strategy used for fitting, respectively.

\subsection{Multi-Resolution 3D Feature Grids}
\label{sec:multires_grid}

\begin{figure}[!htbp]
    \centering
    \includegraphics[width=\linewidth]{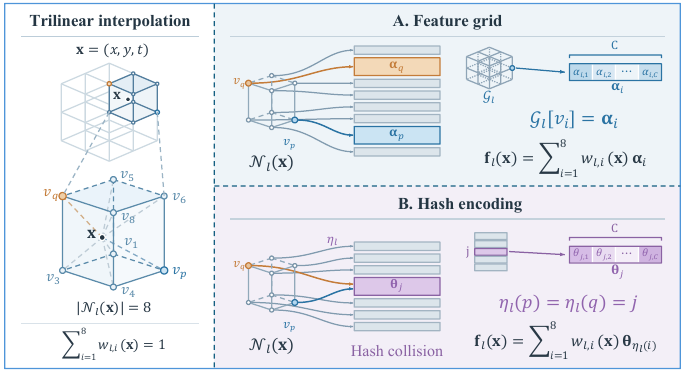}
    \caption{Explicit and hash-constrained 3D feature grids at the same resolution and feature dimension. Both use trilinear interpolation.
\textbf{Feature grid:} Each vertex has an independent feature vector.
\textbf{Hash encoding:} Hash collisions force distinct vertices to share a feature vector, coupling their representations.}
    \label{fig:featuregrid_vs_hash}
\end{figure}

Multi-resolution feature encodings, exemplified by Instant-NGP \cite{instantngp}, combine hash-grid-based learnable local features with a compact neural decoder. As Figure~\ref{fig:featuregrid_vs_hash}(b) shows, hash-grid encoding uses regular interpolation lattices while storing their vertex features in a compact parameter table. At finer levels, where the number of grid vertices exceeds the table capacity, a deterministic spatial hash function maps integer vertex coordinates to table entries. Colliding vertices then share a feature vector, introducing parameter coupling between locations that may not be spatial neighbors.

Compact hash tables can be particularly useful when observations are concentrated in a small part of the spatial domain. In our setting, however, video frames provide dense observations over a regular spatiotemporal domain. As Figure~\ref{fig:featuregrid_vs_hash}(a) shows, we therefore use explicit 3D feature grids that assign an independent feature vector to every vertex, eliminating parameter sharing caused by hash collisions. Compared with a collision-prone hash table at the same lattice resolution and feature dimension, this independent parameterization requires slightly more storage.

Without loss of generality, let $T,H,W>1$. The pixel at temporal index $k$, row $j$, and column $i$ is assigned the normalized coordinate
\begin{equation}
\mathbf{x}_{kji}=\left(\frac{2i}{W-1}-1,\frac{2j}{H-1}-1,\frac{2k}{T-1}-1\right)^\top,
\label{eq:video_coordinates}
\end{equation}
where indices start at zero. This convention treats the observed frames as uniformly spaced in normalized time.

We use $L$ learnable grids with $C$ feature channels per vertex. At each level, the implementation uses the same number of vertices along all three normalized axes:
\begin{equation}
\mathcal{G}_l\in\mathbb{R}^{C\times R_l\times R_l\times R_l},\qquad l=1,\ldots,L.
\label{eq:grid_tensor}
\end{equation}
The tensor axes after the channel dimension correspond to time, height, and width, respectively. Here, $R_l$ counts vertices rather than cells, so each axis contains $R_l-1$ cells with spacing
\begin{equation}
h_l=2/(R_l-1).
\label{eq:grid_spacing}
\end{equation}
This spacing is defined in normalized coordinates and does not imply equal physical scales in space and time.

For a query $\mathbf{x}$, let $\mathcal{N}_l(\mathbf{x})$ denote the eight vertices of a grid cell, as illustrated in Figure~\ref{fig:featuregrid_vs_hash}. Let $w_{l,\mathbf{v}}(\mathbf{x})$ denote the corresponding trilinear weights. The interpolated feature is
\begin{equation}
\mathbf{f}_l(\mathbf{x})=\sum\nolimits_{\mathbf{v}\in\mathcal{N}_l(\mathbf{x})}w_{l,\mathbf{v}}(\mathbf{x})\,\mathcal{G}_l[\mathbf{v}].
\label{eq:explicit_grid_query}
\end{equation}
Within the domain, these trilinear weights are nonnegative and sum to one. Shared vertex features ensure that adjacent cells agree on their common faces, yielding a continuous, piecewise trilinear feature field. These features form $\mathbf{z}(\mathbf{x})$ in Eq.~\eqref{eq:multires_feature}, with a concatenated dimension of $LC=8$ in the main model. Different grid resolutions provide local features with different support sizes. Positional encodings such as the one used in NeRF map coordinates to discrete Fourier frequency bands. In contrast, 3D feature grids provide a multiscale representation through grid granularity and spatial support, without explicitly assigning features to discrete Fourier components.

The parameter count and per-query computational cost of the feature grids are
\begin{equation}
P_{\mathrm{grid}}=C\sum\nolimits_{l=1}^{L}R_l^3~~\text{and}~~
\operatorname{Cost}_{\mathrm{grid}}=\mathcal{O}(LC),
\label{eq:grid_parameters}
\end{equation}
respectively, where $L$ is the number of resolution levels, $R_l$ denotes the number of grid vertices along each dimension at level $l$, and $C$ is the feature dimension per vertex.

\noindent\textbf{Remark 1 (Expressive ability of feature grids).}
Figure~\ref{fig:featuregrid_vs_hash} illustrates the distinction between independent vertex features and hash-induced coefficient sharing. Let $M_l=R_l^3$ and $\{N_{l,i}\}_{i=1}^{M_l}$ be the linearly independent trilinear nodal basis on the level-$l$ grid. The explicit feature space is
\begin{equation}
\mathcal{U}_l^{\mathrm{grid}}=\left\{\sum\nolimits_{i=1}^{M_l}\boldsymbol{\alpha}_iN_{l,i}:\boldsymbol{\alpha}_i\in\mathbb{R}^{C}\right\},
\label{eq:explicit_space}
\end{equation}
where $\boldsymbol{\alpha}_i=\mathcal{G}_l[\mathbf{v}_i]$ denotes the independently learnable feature vector at vertex $\mathbf{v}_i$. For the same vertices, interpolation basis, and feature dimension, fix a hash assignment $\eta_l:\{1,\ldots,M_l\}\to\{1,\ldots,K_l\}$. The corresponding hash feature space is
\begin{equation}
\mathcal{U}_l^{\mathrm{hash}}=
\left\{
\sum_{i=1}^{M_l}\boldsymbol{\theta}_{\eta_l(i)}N_{l,i}:
\boldsymbol{\theta}_j\in\mathbb{R}^{C},\
1\leq j\leq K_l
\right\}.
\label{eq:hash_space}
\end{equation}

By setting $\boldsymbol{\alpha}_i=\boldsymbol{\theta}_{\eta_l(i)}$ for every vertex, any hash-representable field can be expressed in the explicit-grid space, leading to $\mathcal{U}_l^{\mathrm{hash}}\subseteq\mathcal{U}_l^{\mathrm{grid}}$. Moreover, if $\eta_l$ is non-injective, this inclusion is strict. In particular, two distinct vertices $p$ and $q$ with $\eta_l(p)=\eta_l(q)$ must share the same feature coefficient in the hash space. However, the independently parameterized grid permits $\boldsymbol{\alpha}_p\neq\boldsymbol{\alpha}_q$. Since the nodal basis functions are linearly independent, such a feature field cannot be represented by the hash space. Thus, more rigorously, $\mathcal{U}_l^{\mathrm{hash}}\subsetneq\mathcal{U}_l^{\mathrm{grid}}$ whenever hash collisions occur.

Both spaces are finite-dimensional closed subspaces of $L_2(\Omega;\mathbb{R}^{C})$. Let $\Pi_l^{\mathrm{grid}}$ and $\Pi_l^{\mathrm{hash}}$ denote their orthogonal projections. For any target feature field $\mathbf{F}\in L_2(\Omega;\mathbb{R}^{C})$, the hash-space projection residual splits into the explicit-grid projection residual and the difference between the two projections:
\begin{equation}
\begin{aligned}
\|\mathbf{F}-\Pi_l^{\mathrm{hash}}\mathbf{F}\|_{L_2}^2
={}&\|\mathbf{F}-\Pi_l^{\mathrm{grid}}\mathbf{F}\|_{L_2}^2\\
&+\|\Pi_l^{\mathrm{grid}}\mathbf{F}
-\Pi_l^{\mathrm{hash}}\mathbf{F}\|_{L_2}^2,
\end{aligned}
\label{eq:projection_decomposition}
\end{equation}
where the cross term vanishes because $\mathbf{F}-\Pi_l^{\mathrm{grid}}\mathbf{F}$ is orthogonal to $\mathcal{U}_l^{\mathrm{grid}}$, which contains $\Pi_l^{\mathrm{grid}}\mathbf{F}-\Pi_l^{\mathrm{hash}}\mathbf{F}$. Consequently,
\begin{equation}
\|\mathbf{F}-\Pi_l^{\mathrm{grid}}\mathbf{F}\|_{L_2}
\leq
\|\mathbf{F}-\Pi_l^{\mathrm{hash}}\mathbf{F}\|_{L_2},
\label{eq:grid_hash_error_order}
\end{equation}
which suggests that per Eqs. \eqref{eq:explicit_space} and \eqref{eq:hash_space} (the lattice and feature dimensions are fixed), the feature grid representation achieves a feature-field approximation error less than or equal to that of hash-grid encoding by eliminating hash-induced coefficient sharing. Theoretically, the hash-grid encoding can involve higher lattice and feature dimensions to improve the expressive ability; however, it will lead to larger storage.

\subsection{QG: Quadratically Gated Decoder}
\label{sec:QG_decoder}

We map the concatenated grid features to RGB values using a compact decoder with multiplicative gating. Each QG layer multiplies an affine content projection by a SiLU-activated gating projection, where $\operatorname{SiLU}(a)=a(1+e^{-a})^{-1}$ \cite{elfwing2017silu}. This operation follows the SwiGLU form with the Swish parameter \(\beta=1\) \cite{shazeer2020glu}. For an input feature vector $\mathbf{z}\in\mathbb{R}^{d}$, the layer produces $\mathbf{y}\in\mathbb{R}^{M}$ according to
\begin{equation}
\mathbf{y}
=
(\mathbf{W}_{\mathrm{L}}\mathbf{z}+\mathbf{b}_{\mathrm{L}})
\odot
s(\mathbf{W}_{\mathrm{G}}\mathbf{z}+\mathbf{b}_{\mathrm{G}}),
\label{eq:QG_layer}
\end{equation}
where $s=\operatorname{SiLU}$ is applied elementwise, with $\mathbf{W}_{\mathrm{L}},\mathbf{W}_{\mathrm{G}}\in\mathbb{R}^{M\times d}$, and $\mathbf{b}_{\mathrm{L}},\mathbf{b}_{\mathrm{G}}\in\mathbb{R}^{M}$. Since SiLU is not restricted to $[0,1]$, the gate can both amplify and reverse the sign of the content signal rather than merely attenuating it. The hidden layers $g_1$ and $g_2$ in Eq.~\eqref{eq:multires_feature} use this operation, with input dimensions $L\cdot C$ and $M$, respectively, and share an output width $M=64$ in the main configuration. The final projection has $\mathbf{W}_{\mathrm{out}}\in\mathbb{R}^{3\times M}$ and $\mathbf{b}_{\mathrm{out}}\in\mathbb{R}^{3}$, followed by an elementwise sigmoid to produce RGB values. This gated decoder models multiplicative interactions between learned projections of the interpolated grid features through compact affine projections rather than dense quadratic weight matrices.

\noindent\textbf{Local second-order interactions.}
For the \(m\)-th QG neuron, write $y_m(\mathbf{z})=u_m s(v_m)$, where $u_m=\mathbf{p}_m^\top\mathbf{z}+b_{\mathrm{L},m}$ and $v_m=\mathbf{q}_m^\top\mathbf{z}+b_{\mathrm{G},m}$. Here, \(\mathbf p_m^\top\) and \(\mathbf q_m^\top\) denote the \(m\)-th rows of \(\mathbf W_{\mathrm L}\) and \(\mathbf W_{\mathrm G}\), respectively. Applying the product and chain rules yields the Hessian
\begin{equation}
\begin{aligned}
\nabla_{\mathbf{z}}^2 y_m
={}&s'(v_m)(\mathbf{p}_m\mathbf{q}_m^\top+\mathbf{q}_m\mathbf{p}_m^\top)\\
&+u_m s''(v_m)\mathbf{q}_m\mathbf{q}_m^\top.
\end{aligned}
\label{eq:QG_hessian}
\end{equation}

Since the Hessian's range lies in $\operatorname{span}\{\mathbf{p}_m,\mathbf{q}_m\}$, its rank is at most $\min\{d,2\}$, revealing the low-rank structure of the neuron's local second-order interactions. This multiplicative structure allows the gate along \(\mathbf q_m\) to modulate the content response along \(\mathbf p_m\), capturing dependencies among interpolated grid features without a full quadratic weight matrix. Such interactions help the decoder model complex spatiotemporal variations. This pointwise rank bound applies to the Hessian of an individual QG neuron with respect to its immediate input, and does not generally extend to the Hessian of the full decoder. Although each QG neuron exhibits local second-order interactions, its SiLU-gated mapping is generally not quadratic in its input.

We next analyze the parameter count and computational complexity of QG layers. For a layer with input dimension $d$ and output width $m$, these are
\begin{equation}
P_{\mathrm{QG}}=2m(d+1)~~\text{and}~~
\operatorname{Cost}_{\mathrm{QG}}=\mathcal{O}(md),
\label{eq:QG_layer_cost}
\end{equation}
respectively. For fixed \(m\), the QG parameter count scales linearly with \(d\), whereas unrestricted quadratic forms for \(m\) output units require \(\mathcal{O}(md^2)\) parameters. The two affine projections require only slightly more computation than a single projection with the same input and output dimensions. With two QG layers of hidden width $m$ and a final RGB output layer, the decoder contains
\begin{equation}
P_{\mathrm{dec}}=2(LCm+m)+2(m^2+m)+(3m+3).
\label{eq:decoder_parameters}
\end{equation}

For the main configuration with $L\cdot C=8$ and $m=64$, Eq.~\eqref{eq:decoder_parameters} gives $P_{\mathrm{dec}}=9{,}667$. Compared with a same-width affine layer, multiplicative gating increases the computational cost per optimization step but introduces low-rank second-order interactions that accelerate the decoder to reach a target reconstruction quality in fewer updates. This improvement can offset the higher cost per step, leading to better reconstruction quality within a fixed training-time budget.

\subsection{Error-Aware Sampling}
\label{sec:sampling_policies}

\begin{figure}[!htbp]
\centering
\includegraphics[width=0.9\linewidth]{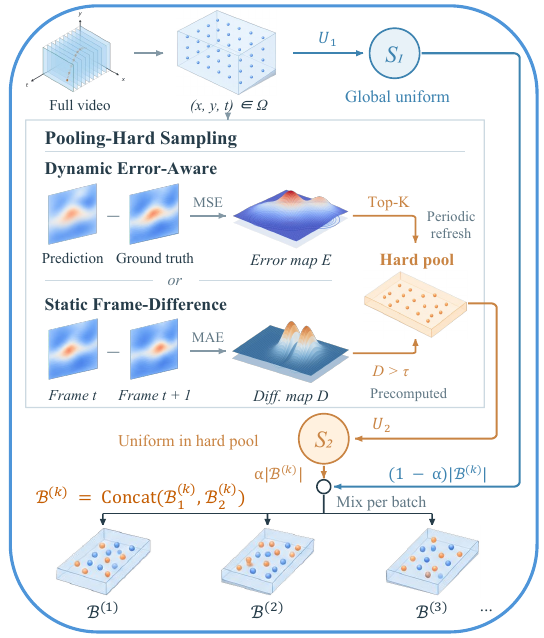}
\caption{Error-aware coordinate sampling. Coordinates from two branches are combined into a batch for training. One branch samples uniformly from the entire observed sequence, while the other draws from a pool of difficult coordinates constructed using reconstruction errors (dynamic sampling) or differences between adjacent frames (static sampling).}
\label{fig:pool_hard}
\end{figure}

Given this representation, we next consider how to select training coordinates to speed up fitting. Let $\mathcal{X}_{\mathrm{obs}}=\{\mathbf{x}_{kji}\mid 0\leq k<T,\ 0\leq j<H,\ 0\leq i<W\}$ denote the set of pixel coordinates, where $T$, $H$, and $W$ are the number of video frames, frame height, and frame width, respectively, and $\mathbf{x}_{kji}$ is the normalized spatiotemporal coordinate of the pixel at frame $k$, row $j$, and column $i$. As illustrated in Figure~\ref{fig:pool_hard}, we combine two branches of coordinate sampling: \(\mathcal S_1\) that uniformly samples the entire observed coordinate set \(\mathcal X_{\mathrm{obs}}\); \(\mathcal S_2\) that samples a pool of important candidates. The dynamic variant of \(\mathcal S_2\) constructs the candidate pool from recorded reconstruction errors, while the static variant uses differences between adjacent frames. We retain a uniform sampling component to prevent coordinates outside the prioritized pool from being excluded during training, causing underfitting over these coordinates. We next describe how the dynamic and static sampling strategies construct their respective candidate pools.

\underline{Dynamic Sampling.}
Reconstruction errors provide feedback on where the current representation requires further refinement. We therefore maintain a sequence-wide error map $\mathcal{E}\in\mathbb{R}^{T\times H\times W}$ to guide coordinate selection. For each sampled coordinate \(\mathbf x\), we compute the mean squared error over RGB channels which is
\begin{equation}
\ell(\mathbf{x})=
\frac{1}{3}\sum\nolimits_{c=1}^{3}
\left(
\widehat{\mathbf{V}}_c(\mathbf{x})
-\mathbf{V}_c(\mathbf{x})
\right)^2.
\label{eq:sample_error}
\end{equation}
We update the error map by direct indexing, \(\mathcal E_{kji}\leftarrow\ell(\mathbf x_{kji})\). The map thus records the latest observed reconstruction errors rather than an instantaneous evaluation of the entire video. These errors can reflect fitting difficulties associated with motion, fine detail, or other complex spatiotemporal content. This error-based feedback requires no explicit motion estimation.

At each pool refresh, we select the $N_{\mathrm{pool}}$ coordinates associated with the largest recorded errors:
\begin{equation}
\mathcal{P}_{\mathrm{hard}}=
\left\{
\mathbf{x}_{kji}:
(k,j,i)\in
\operatorname{TopK}(\mathcal{E},N_{\mathrm{pool}})
\right\},
\label{eq:dynamic_pool}
\end{equation}
where $\operatorname{TopK}$ returns index triples. Entries not sampled in the current step retain their initialized or previously recorded values. Assuming $\mathcal{P}_{\mathrm{hard}}\neq\emptyset$, for both sampling variants, the prioritized branch $\mathcal{S}_2$ samples uniformly from the corresponding nonempty candidate pool $\mathcal{P}_{\mathrm{hard}}$, following
\begin{equation}
p_{\mathrm{hard}}(\mathbf{x})=
\begin{cases}
|\mathcal{P}_{\mathrm{hard}}|^{-1},
&\mathbf{x}\in\mathcal{P}_{\mathrm{hard}},\\
0,&\text{otherwise},
\end{cases}
\quad
\mathbf{x}\in\mathcal{X}_{\mathrm{obs}}.
\label{eq:hard_distribution}
\end{equation}
Before the first pool refresh, dynamic sampling uses only the uniform branch to select training coordinates. Reconstruction errors come from the forward pass already used for training, so updating the error map requires no additional forward pass. To limit the cost of pool selection, we refresh the candidate pool every $K_{\mathrm{eval}}=1,000$ steps rather than at each step.

\underline{Static Sampling.}
As a simpler alternative, static sampling constructs a fixed candidate pool from differences between adjacent video frames. Using the discrete indices defined in Eq.~\eqref{eq:video_coordinates}, the mean absolute RGB difference between consecutive frames is
\begin{equation}
D_{kji}=
\frac{1}{3}\sum\nolimits_{c=1}^{3}
\left|
\mathbf{V}_{k+1,j,i,c}
-\mathbf{V}_{k,j,i,c}
\right|
\label{eq:frame_difference}
\end{equation}
for $0\leq k<T-1$, with $D_{T-1,j,i}=D_{T-2,j,i}$ at the final frame. Here, \(\mathbf V_{k,j,i,c}\) denotes the value of color channel \(c\) at frame \(k\), row \(j\), and column \(i\). Applying the threshold $\tau$ to the temporal difference scores yields
\begin{equation}
\mathcal{P}_{\mathrm{hard}}=
\left\{
\mathbf{x}_{kji}\in\mathcal{X}_{\mathrm{obs}}:
D_{kji}>\tau
\right\}.
\label{eq:static_pool}
\end{equation}
\(\mathcal S_2\) samples uniformly from this fixed pool according to Eq.~\eqref{eq:hard_distribution}. Unlike the dynamic variant, static sampling constructs its candidate pool before fitting and keeps it fixed throughout optimization, free from the error-map maintenance and periodic pool refreshes in dynamic sampling. Frame differences capture appearance changes caused by motion, occlusion, or illumination variation, providing a fixed temporal-change prior. This prior reflects where the input changes rather than where the current model struggles, so it may miss regions that are temporally smooth yet difficult to reconstruct.

For both sampling variants, let \(U_1=\mathcal X_{\mathrm{obs}}\) and \(U_2=\mathcal P_{\mathrm{hard}}\) denote the sampling domains of \(\mathcal S_1\) and \(\mathcal S_2\), respectively. Each samples with replacement, and the resulting samples are concatenated without removing duplicate coordinates to form a training batch $\mathcal{B}$:
\begin{equation}
\begin{aligned}
\mathcal{B} &= \operatorname{Concat}(\mathcal{B}_{1},\mathcal{B}_{2}), \\
\mathcal{B}_{1} &\sim \operatorname{Uniform}(U_1, (1-\alpha)|\mathcal{B}|), \\
\mathcal{B}_{2} &\sim \operatorname{Uniform}(U_2, \alpha |\mathcal{B}|),
\end{aligned}
\end{equation}
where \(\alpha\in[0,1)\) is the fraction of the batch sampled from $U_2$, while the remaining fraction $1-\alpha$ is sampled from $U_1$.

\noindent\textbf{Remark 2 (Error-aware sampling without auxiliary prediction).}
Our strategy allocates additional training samples to selected coordinates that require further refinement without an auxiliary prediction network. In the dynamic variant, reconstruction errors from ordinary training directly identify regions that require further refinement, avoiding additional models to predict reconstruction difficulty. Maintaining the error map and refreshing the candidate pool is lighter than training an auxiliary model. Furthermore, combining sequence-wide uniform sampling ensures sufficient coordinate coverage instead of only focusing on difficult samples. Finally, we highlight that although this prediction-free sampling strategy is introduced in implicit neural video representation, it is actually a universal coordinate sampling method, which can be smoothly translated to other INRs and signal domains.

\begin{table*}[!htbp]

\centering
\caption{Comparison of different models at approximately 3M and 5M parameters. Time includes training and quantization for neural methods and encoding for H.265. H.265 uses the medium preset at CRF~28. HNeRV uses 8-bit decoder weights and 6-bit embeddings. FFNeRV uses its 8-bit weight transform. The remaining neural methods use 8-bit quantization. Our variants use chunked decoding with AMP FP16. Bold-faced values indicate the better result between our static and dynamic variants.}
\label{tab:sota_comparison}
\resizebox{\textwidth}{!}{
\renewcommand{\arraystretch}{1.10}
\begin{tabular}{l c | ccccc | ccccc}
\toprule
\rowcolor{headerblue}
\multirow{2}{*}{\textbf{Method}} & \multirow{2}{*}{\textbf{Params (M)}} & \multicolumn{5}{c|}{\textbf{\textit{Beauty} Sequence}} & \multicolumn{5}{c}{\textbf{\textit{Bosphorus} Sequence}} \\
\cmidrule(lr){3-7} \cmidrule(lr){8-12}
& & PSNR$\uparrow$ & MS-SSIM$\uparrow$ & Time (s)$\downarrow$ & BPP$\downarrow$ & FPS$\uparrow$ & PSNR$\uparrow$ & MS-SSIM$\uparrow$ & Time (s)$\downarrow$ & BPP$\downarrow$ & FPS$\uparrow$ \\
\midrule
H.265 \cite{sullivan2012overview} & 0.000 & 33.30 & 0.8943 & 10.23 & 0.01849 & 143.03 & 36.58 & 0.9676 & 11.61 & 0.02054 & 150.91 \\
\midrule
\multicolumn{12}{c}{\cellcolor{gray!15}\textbf{Small-Model Regime} ($\sim$3M Parameters)} \\
\midrule
HNeRV \cite{hnerv} & 3.001 & 31.69 & 0.8686 & 5378.78 & 0.01478 & 366.26 & 29.41 & 0.8758 & 4796.08 & 0.01474 & 360.52 \\
HiNeRV \cite{hinerv2023} & 3.173 & 33.36 & 0.8913 & 7345.86 & 0.01899 & 16.91 & 35.87 & 0.9629 & 7647.15 & 0.01919 & 16.92 \\
FFNeRV \cite{lee2023ffnerv} & 3.033 & 32.38 & 0.8809 & 7265.61 & 0.01351 & 184.62 & 31.89 & 0.9245 & 6698.77 & 0.01398 & 181.83 \\
RNeRV \cite{gwilliam2025design} & 2.993 & 32.24 & 0.8716 & 958.65 & 0.01825 & 139.11 & 31.39 & 0.8933 & 941.59 & 0.01831 & 140.39 \\
Instant-NGP \cite{instantngp} & 3.182 & 30.65 & 0.8463 & 203.68 & 0.01459 & 18.37 & 30.17 & 0.8612 & 204.80 & 0.01468 & 18.35 \\
\midrule
\rowcolor{rowours} \textbf{Ours (Static)} & 3.029 & 31.98 & 0.8766 & \cellcolor{bestcell}{\textbf{33.46}} & 0.01916 & \cellcolor{bestcell}{\textbf{433.74}} & 32.22 & 0.9265 & \cellcolor{bestcell}{\textbf{33.99}} & \cellcolor{bestcell}{\textbf{0.01862}} & 411.48 \\
\rowcolor{rowours} \textbf{Ours (Dynamic)} & 3.029 & \cellcolor{bestcell}{\textbf{32.04}} & \cellcolor{bestcell}{\textbf{0.8774}} & 33.94 & \cellcolor{bestcell}{\textbf{0.01911}} & 418.27 & \cellcolor{bestcell}{\textbf{32.48}} & \cellcolor{bestcell}{\textbf{0.9319}} & 36.99 & 0.01878 & \cellcolor{bestcell}{\textbf{423.08}} \\
\midrule
\multicolumn{12}{c}{\cellcolor{gray!15}\textbf{Standard-Model Regime} ($\sim$5M Parameters)} \\
\midrule
HNeRV \cite{hnerv} & 4.974 & 33.00 & 0.8897 & 6692.22 & 0.02651 & 123.56 & 32.03 & 0.9200 & 5922.10 & 0.02653 & 122.32 \\
HiNeRV \cite{hinerv2023} & 4.976 & 33.57 & 0.8947 & 9167.19 & 0.02946 & 14.52 & 36.66 & 0.9691 & 8516.01 & 0.02971 & 14.54 \\
FFNeRV \cite{lee2023ffnerv} & 5.048 & 32.92 & 0.8890 & 7281.37 & 0.02192 & 162.40 & 33.68 & 0.9440 & 6717.86 & 0.02288 & 157.10 \\
RNeRV \cite{gwilliam2025design} & 5.005 & 32.76 & 0.8812 & 810.79 & 0.03015 & 166.32 & 33.00 & 0.9239 & 807.57 & 0.03026 & 173.26 \\
Instant-NGP \cite{instantngp} & 5.019 & 30.82 & 0.8486 & 264.39 & 0.02473 & 12.27 & 30.36 & 0.8647 & 263.63 & 0.02489 & 12.35 \\
\midrule
\rowcolor{rowours} \textbf{Ours (Static)} & 5.058 & 32.47 & \cellcolor{bestcell}{\textbf{0.8840}} & 48.52 & \cellcolor{bestcell}{\textbf{0.03173}} & 388.18 & 33.14 & 0.9402 & \cellcolor{bestcell}{\textbf{34.92}} & \cellcolor{bestcell}{\textbf{0.03075}} & \cellcolor{bestcell}{\textbf{386.38}} \\
\rowcolor{rowours} \textbf{Ours (Dynamic)} & 5.058 & \cellcolor{bestcell}{\textbf{32.48}} & 0.8839 & \cellcolor{bestcell}{\textbf{42.36}} & 0.03174 & \cellcolor{bestcell}{\textbf{393.41}} & \cellcolor{bestcell}{\textbf{33.30}} & \cellcolor{bestcell}{\textbf{0.9440}} & 36.77 & 0.03100 & 374.02 \\
\bottomrule
\end{tabular}
}
\end{table*}

% Compact float spacing within the experimental sections, without scaling content.
\setlength{\textfloatsep}{4pt plus 1pt minus 1pt}
\setlength{\floatsep}{4pt plus 1pt minus 1pt}
\setlength{\intextsep}{3pt plus 1pt minus 1pt}
\setlength{\dbltextfloatsep}{6pt plus 1pt minus 1pt}
\section{Performance Evaluation and Comparative Analysis}
\label{sec:comparative_evaluation}

We first evaluate Q-GAS as a complete video representation method, comparing its reconstruction quality, storage requirements, fitting time, and decoding throughput with those of existing methods. These metrics involve inherent trade-offs: gains in reconstruction quality often require more storage or computation. To evaluate these trade-offs under model-size and training-time constraints, we use matched-size and fixed-time benchmarks in our experiments.

\begin{figure}[!htbp]

        \centering
    \includegraphics[width=\linewidth,trim=0 4bp 0 3bp,clip]{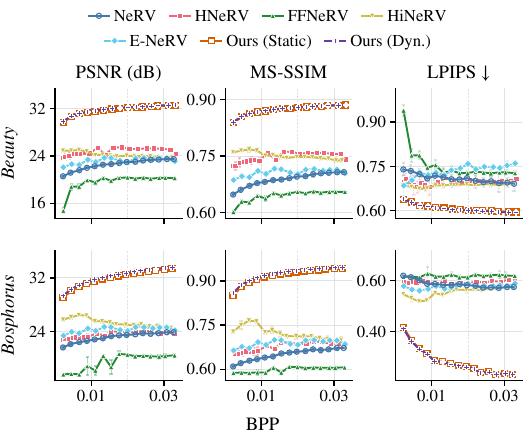}
    \caption{Rate--distortion performance after 90\,s of training. Rows: \textit{Beauty} and \textit{Bosphorus}. Columns: PSNR, MS-SSIM, and LPIPS. Quality and BPP use the same 8-bit quantized model. Points show means, with horizontal and vertical error bars indicating two-sided 95\% Student's $t$ confidence intervals.}
    \label{fig:beauty_bosphorus_rate_distortion_90s}

\end{figure}

\begin{figure}[!htbp]
    \centering
    \includegraphics[width=\linewidth,trim=0 2bp 0 3bp,clip]{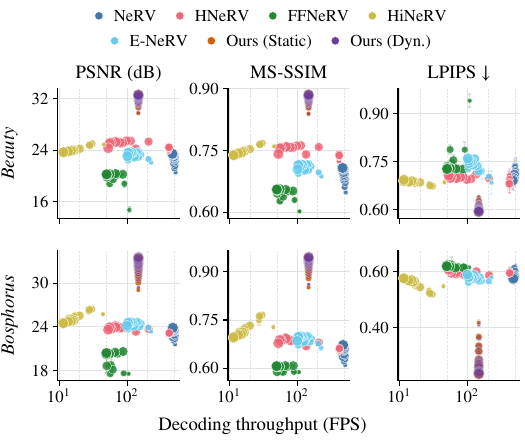}
    \caption{{Quality versus decoding throughput after 90\,s of training. Panel order and statistical summaries follow Figure~\ref{fig:beauty_bosphorus_rate_distortion_90s}. The FPS axis is logarithmic, and size of marker area indicates mean BPP. Q-GAS methods use uncompiled FP32 full-frame decoding.}}
    \label{fig:beauty_bosphorus_quality_speed_90s}
\end{figure}

\begin{figure}[!htbp]
    \centering
    \includegraphics[width=\linewidth,trim=0 2bp 0 3bp,clip]{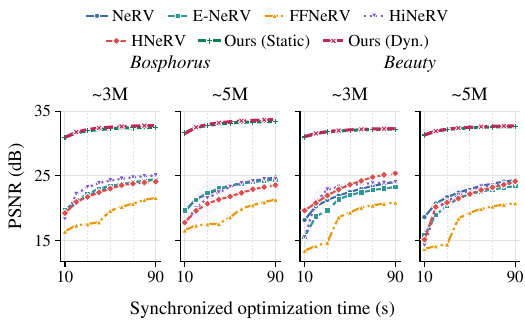}
    \caption{Mean frame-wise PSNR after 8-bit quantization versus synchronized optimization time for approximately 3M- and 5M-parameter models. Each budget is evaluated independently over 5 runs. Points show means, and shaded bands show two-sided 95\% Student's $t$ confidence intervals.}
    \label{fig:optimization_compute_convergence}
\end{figure}

\noindent\textbf{Implementation Details}. For the matched-size comparison in Table~\ref{tab:sota_comparison}, we evaluate approximately 3M- and 5M-parameter models against HNeRV \cite{hnerv}, FFNeRV \cite{lee2023ffnerv}, RNeRV \cite{gwilliam2025design}, HiNeRV \cite{hinerv2023}, Instant-NGP \cite{instantngp}, and include H.265/HEVC \cite{sullivan2012overview} as a codec reference. The NeRV-family baselines are adapted from their public implementations. RNeRV uses the public VINRB implementation \cite{gwilliam2025design} with channel widths adjusted to match the target model capacities. For the fixed-budget experiments in Figures~\ref{fig:beauty_bosphorus_rate_distortion_90s}--\ref{fig:optimization_compute_convergence}, we use adapted baseline pipelines, including implementations built on VINRB. Within each benchmark, shared data interfaces and evaluation procedures allow comparison on the same videos under the specified constraints, including common 8-bit quantization for the fixed-budget experiments. Our Instant-NGP baseline uses a PyTorch hash-grid implementation, so its throughput is not representative of the original fused CUDA implementation \cite{instantngp}.

During training, we jointly optimize the grid coefficients and decoder parameters of Q-GAS by minimizing the reconstruction loss in Eq.~\eqref{eq:reconstruction_loss} using Adam \cite{kingma2014adam} with a OneCycle learning-rate schedule \cite{smith2019super}. Q-GAS uses the same grids and decoder with either dynamic or static sampling. Both sampling configurations and the Instant-NGP baseline use 16,000 updates, whereas the NeRV-family models use 100 epochs. The matched-size comparison therefore follows the respective training protocols of the evaluated implementations rather than enforcing an identical number of optimization steps.

We measure reconstruction quality using PSNR and MS-SSIM \cite{wang2003multiscale}. The 90\,s experiments also evaluate SSIM \cite{wang2004image} and learned perceptual similarity (LPIPS) \cite{zhang2018unreasonable}. The mean LPIPS-AlexNet v0.1 over all full-resolution frames is reported, where lower values indicate greater perceptual similarity. To account for actual storage rather than parameter count alone, we report the size of each quantized neural archive or H.265 bitstream in bits per pixel (BPP):
\begin{equation}
\mathrm{BPP} = 8\,S_{\mathrm{file}}/(T\times H\times W),
\label{eq:bitrate_bpp}
\end{equation}
where $S_{\mathrm{file}}$ denotes the file size in bytes, including quantization metadata and container overhead. $T$, $H$, and $W$ denote the number of frames in the complete sequence, frame height, and frame width, respectively. The factor $8$ converts bytes to bits, while $T\times H\times W$ represents the total number of pixels across all frames, regardless of the number of color channels.

For equally sized frames with RGB values in $[0,1]$, let $m_t$ denote the MSE over all RGB values in frame $t$, and $\overline{m}=T^{-1}\sum_t m_t$ denote the global video MSE. Table~\ref{tab:sota_comparison} reports $10\log_{10}(1/\overline{m})$, which gives poorly reconstructed frames greater influence than averaging frame-wise PSNR. The fixed-budget curves instead report mean frame-wise PSNR, $T^{-1}\sum_t10\log_{10}(1/m_t)$, alongside mean frame-wise SSIM and MS-SSIM. Because these two PSNR aggregation methods generally yield different values, we compare absolute PSNR scores only under the same evaluation protocol.

All experiments use the complete 600-frame \textit{Beauty}, \textit{Bosphorus}, and \textit{Jockey} sequences from the UVG dataset \cite{meratnia2020uvg}, using their $1,920\times1,080$ versions. We implement the neural methods in PyTorch~\cite{paszke2019pytorch}. Each run uses a single NVIDIA RTX PRO 6000 Blackwell Server Edition GPU.

\noindent\textbf{Comparison at Matched Model Size.} As Table~\ref{tab:sota_comparison} shows, Q-GAS with dynamic sampling achieves competitive reconstruction quality at comparable model sizes while fitting and decoding rapidly. On both \textit{Beauty} and \textit{Bosphorus}, training and quantization together take less than 50\,s, while mixed-precision decoding exceeds 370 FPS. Compared with the evaluated Instant-NGP implementation, Q-GAS achieves higher PSNR, completes training and quantization 5.5--7.2$\times$ faster, and decodes at 20--32$\times$ its throughput. RNeRV requires approximately 19--28$\times$ the training and quantization time of Q-GAS. Although HiNeRV achieves higher PSNR, it takes substantially longer to fit and decodes more slowly. These results highlight Q-GAS's efficiency in rapid fitting and high-throughput video decoding.

Feature grids assign independent coefficients to lattice vertices, so increasing grid resolution provides more local capacity but requires more storage. This helps explain why larger Q-GAS models improve reconstruction fidelity, particularly on \textit{Bosphorus}. Thus, Q-GAS offers a favorable quality–cost trade-off rather than a universal compression advantage.

\noindent\textbf{Quality under a Fixed Training Budget.}
The matched-size results leave open what reconstruction quality each method can achieve within the same training time. We therefore limit training to 90\,s and further vary model size across a range of storage costs. This tests both how quickly a representation can be fitted and whether increased model size leads to better reconstructions within the limited training time.

We time the training phase after loading the data and setting up the model and optimizer. Any initialization performed within training, such as FFNeRV's key-frame initialization, remains included. Training stops at an optimization-step boundary once the measured time reaches 90\,s. The measured time includes all operations during fitting but excludes subsequent quantization, archive construction, and final evaluation. For each method, reconstruction quality and BPP are then evaluated using its symmetrically quantized 8-bit archive, with packed frame embeddings included in the HNeRV archive. We use automatic mixed precision (AMP) to accelerate Q-GAS fitting, together with a OneCycle learning-rate schedule configured for the training budget. The NeRV-family baselines retain FP32 training and their configured schedules. The comparison then measures the performance of the implemented training pipelines under a common time budget.

We first compare methods at comparable BPP. As shown in Figure~\ref{fig:beauty_bosphorus_rate_distortion_90s}, Q-GAS with dynamic sampling achieves higher PSNR and MS-SSIM and lower LPIPS than the five NeRV-family baselines at comparable BPP across the overlapping measured ranges. Despite strong reconstruction quality in the matched-size comparison, HiNeRV does not maintain the same relative advantage under the 90\,s budget. This difference highlights the importance of considering both model storage and available fitting time when evaluating video representations.

Increasing grid resolution generally improves Q-GAS's PSNR and MS-SSIM while reducing LPIPS on both sequences. In contrast, several NeRV-family baselines show non-monotonic changes in reconstruction quality as model capacity increases, particularly HiNeRV. Under the same 90\,s training budget, larger models may not be sufficiently optimized to translate their additional representational capacity into better reconstructions. This may explain why increasing model size does not consistently improve their reconstruction quality.

\noindent\textbf{Quality--Throughput Trade-offs.} Using the same models fitted under the 90\,s budget, we next examine the trade-off between reconstruction quality and decoding throughput. For this comparison, decoding time includes all operations required to reconstruct frames but excludes initial model loading, source-frame input/output (I/O), and quality evaluation. As shown in Figure~\ref{fig:beauty_bosphorus_quality_speed_90s}, Q-GAS with dynamic sampling combines high PSNR and MS-SSIM with low LPIPS at approximately 146 FPS using FP32 decoding. Although NeRV and some HNeRV configurations decode faster, they yield lower reconstruction accuracy and poorer perceptual similarity. Q-GAS thus offers a favorable balance between reconstruction quality and decoding speed for high-quality playback.

Over the tested range, increasing grid resolution improves reconstruction quality with little change in decoding throughput. To reconstruct the RGB value at a given spatiotemporal coordinate, Q-GAS interpolates local features from each grid level and passes them to the decoder. Finer grids store more coefficients, but the number of grid levels, the number of vertices accessed per level, the feature dimension, and the decoder architecture remain unchanged. This helps explain why additional grid capacity improves quality with little change in decoding speed. Similarly, dynamic and static sampling yield comparable decoding speeds because they affect training while leaving the reconstruction process unchanged.

\Needspace{3\baselineskip}
\noindent\textbf{Quality across Optimization Budgets.} The preceding comparisons evaluate reconstruction quality and decoding throughput after 90\,s training. We next examine how early the quality advantage emerges by varying the optimization-time budget for models with approximately 3M and 5M parameters. Figure~\ref{fig:optimization_compute_convergence} reports synchronized optimization time, excluding data loading and host-to-device transfer overhead.

\begin{figure*}[!htbp]
    \centering
    \includegraphics[width=\linewidth]{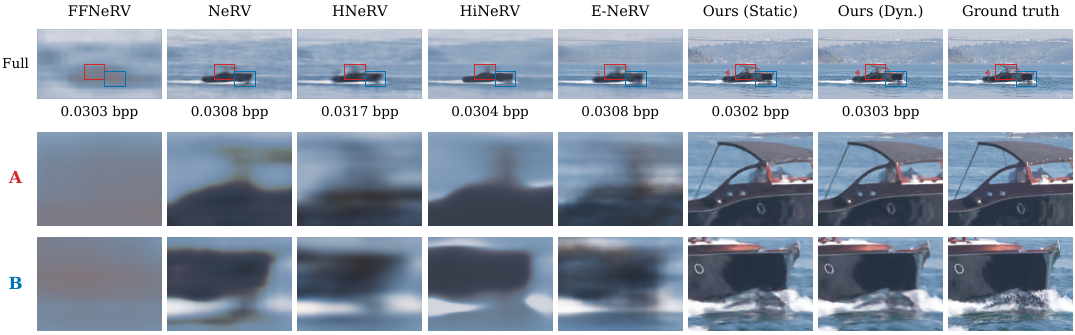}
    \caption{Full-frame reconstructions of \textit{Bosphorus} (frame 382) and corresponding enlarged regions after 90\,s training, using 8-bit quantized models. Red and blue boxes mark regions A and B, which are enlarged in the second and third rows, respectively. For each method, we select the configuration with the lowest mean BPP at or above 0.03. Five baselines are then arranged from left to right in ascending order of full-frame PSNR.}
    \label{fig:bosphorus_90s_q8}
\end{figure*}

Across all four settings, Q-GAS with dynamic sampling achieves higher PSNR after just 10\,s than any evaluated NeRV-family baseline achieves after 90\,s. Q-GAS then approaches a plateau, while the baselines continue to improve even near the end of the measured interval. These results indicate that Q-GAS reaches higher reconstruction quality substantially earlier under the tested conditions, with the advantage persisting even when data loading and transfer overhead are excluded.

\noindent\textbf{Visual Comparison.} Figure~\ref{fig:bosphorus_90s_q8} compares reconstruction quality at similar storage rates after 90\,s training. In region A, Q-GAS with dynamic sampling retains the thin canopy support and curved cabin boundary, which become blurred or indistinct in the baseline reconstructions. Region B shows a similar advantage along the bow and porthole, but also reveals a limitation: fine water spray remains smoothed relative to the ground truth. Q-GAS with static sampling recovers similar visible structures in these regions. Overall, our approach reconstructs scene structures more clearly within the specified training budget, although fine high-frequency textures remain an area for improvement.

\section{Experimental Analysis}
\label{sec:experiments}

Following the main benchmark comparisons, we analyze how decoder design and coordinate sampling contribute to Q-GAS. We first compare compact decoder alternatives for rapid per-video fitting, then isolate the effect of QG in the full architecture and examine full-sequence fitting trajectories. We subsequently study uniform, dynamic, and static sampling, including their interactions with the selected decoder.

\begin{table}[!htbp]
\centering
\caption{Comparison of compact decoders in models with approximately 11.40M parameters. Results are reported as means $\pm$ sample standard deviations over completed runs. The best mean for each metric within each sequence is bold-faced.}
\label{tab:decoder_repeats}
\footnotesize
\setlength{\tabcolsep}{3pt}
\renewcommand{\arraystretch}{1.05}
\begin{tabular}{lccc}
\toprule
\rowcolor{headerblue}
Decoder & PSNR (dB)$\uparrow$ & Time (s)$\downarrow$ & GPU FPS$\uparrow$ \\
\midrule
\multicolumn{4}{c}{\cellcolor{gray!15}\textbf{\textit{Beauty}}} \\
\midrule
Linear & $32.557\pm0.069$ & $34.559\pm0.636$ & $\mathbf{1486.6}\pm2.1$ \\
FAN & $32.673\pm0.112$ & $20.353\pm0.625$ & $631.7\pm0.2$ \\
\rowcolor{rowours}
\textbf{QG} & $\mathbf{32.737}\pm0.039$ & $\mathbf{18.270}\pm0.354$ & $898.9\pm1.5$ \\
\midrule
\multicolumn{4}{c}{\cellcolor{gray!15}\textbf{\textit{Bosphorus}}} \\
\midrule
Linear & $33.339\pm0.180$ & $34.736\pm1.162$ & $\mathbf{1484.7}\pm4.7$ \\
FAN & $\mathbf{33.763}\pm0.118$ & $20.747\pm0.359$ & $629.8\pm5.0$ \\
\rowcolor{rowours}
\textbf{QG} & $33.714\pm0.156$ & $\mathbf{18.616}\pm0.907$ & $896.4\pm7.3$ \\
\bottomrule
\end{tabular}
\end{table}

\noindent\textbf{Decoder Choice for Rapid Fitting.}
We first compare linear neurons, the quadratic neurons introduced by Fan et al. (FAN)~\cite{fan2018new}, and our QG neurons in a compact fully connected decoder that maps interpolated grid features to RGB values. For FAN, our implementation applies SiLU to one branch before multiplication. To examine whether neuron choice can improve fitting efficiency with a compact decoder, we use two eight-channel feature grids and a decoder with two hidden layers of eight units each, leaving most parameters in the grids. Each candidate uses its designated neuron type in both hidden layers and the RGB output layer, with ReLU applied after each hidden layer. This is a separate compact comparison, whereas the main architecture uses an affine RGB output. All candidates use uniform sampling and FP16 mixed precision, while retaining their respective initialization schemes, predefined learning rates, and training schedules. After reloading the FP16 checkpoints, we evaluate mean frame-wise PSNR. To characterize computational efficiency, we report optimization time excluding compilation and setup, together with warmed-up GPU forward throughput.

Table~\ref{tab:decoder_repeats} shows that the fastest decoder at inference need not yield the shortest fitting time. Linear has the highest forward throughput, yet QG achieves higher mean PSNR with less optimization time on both sequences under their respective schedules. Although each QG update is more computationally expensive, the smaller number of updates in its training schedule leads to lower total optimization time. FAN's \textit{Bosphorus} statistics exclude one run that failed numerically. Compared with FAN, QG attains slightly higher mean PSNR on Beauty but slightly lower mean PSNR on Bosphorus. Nevertheless, QG requires less optimization time and achieves higher decoding throughput than FAN on both sequences. This sequence-dependent tradeoff supports QG as a practical choice when both rapid fitting and efficient reconstruction matter.

\begin{figure}[!t]
\centering
\includegraphics[width=0.9\columnwidth,trim=0 3bp 0 1bp,clip]{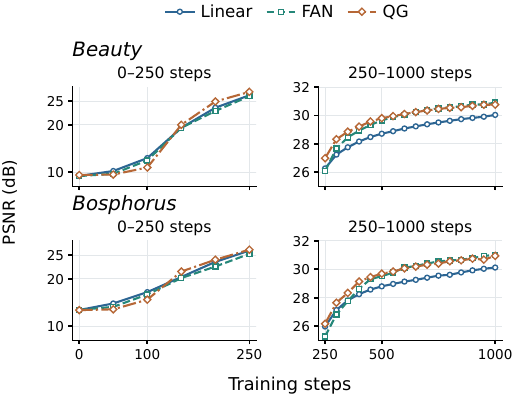}
\caption{Early fitting performance of the decoders in Table~\ref{tab:decoder_repeats} over the first 1,000 updates. Markers show mean frame-wise PSNR every 50 updates. The two training windows use different PSNR ranges to reveal fitting differences.}
\label{fig:decoder_tradeoffs}
\end{figure}

To better understand the fitting-time trade-offs in Table~\ref{tab:decoder_repeats}, Figure~\ref{fig:decoder_tradeoffs} tracks reconstruction quality over the first 1,000 updates. Total fitting time depends on both the cost of each update and the number of updates needed to reach a given quality. Although QG initially trails Linear, it subsequently reaches comparable PSNR in fewer updates on both sequences. Requiring fewer updates can compensate for its higher per-update cost, consistent with the shorter fitting times reported in Table~\ref{tab:decoder_repeats}. FAN gradually reaches similar or slightly higher PSNR, so the quality trajectories alone do not determine which decoder fits fastest. Since this comparison retains the respective training schedules, we next examine QG in the full architecture under a common update budget.

\begin{table}[!t]
\centering
\caption{Decoder comparison in the two-hidden-layer architecture with 64 neurons at each layer, using the same feature grids and uniform sampling. All variants use a common 20,000-step schedule and unquantized FP32 evaluation. FG denotes the 3D feature grid. Green shading highlights the QG configurations. The best result for each metric is bold-faced.}
\label{tab:ablation_comparison}
\footnotesize
\setlength{\tabcolsep}{2pt}
\renewcommand{\arraystretch}{1.05}
\begin{tabularx}{0.96\columnwidth}{>{\raggedright\arraybackslash}Xc|cccc}
\toprule
\rowcolor{headerblue}
Method & Params (M) & Time (s)$\downarrow$ & FPS$\uparrow$ & PSNR (dB)$\uparrow$ & MS-SSIM$\uparrow$ \\
\midrule
\multicolumn{6}{c}{\cellcolor{gray!15}\textbf{\textit{Beauty}}} \\
\midrule
FG + MLP & 5.05 & \textbf{32.48} & \textbf{400.43} & 32.39 & 0.8833 \\
\rowcolor{rowours}
\textbf{FG + QG} & 5.06 & 40.48 & 181.37 & \textbf{32.52} & \textbf{0.8841} \\
\midrule
\multicolumn{6}{c}{\cellcolor{gray!15}\textbf{\textit{Jockey}}} \\
\midrule
FG + MLP & 5.05 & \textbf{32.91} & \textbf{407.66} & 28.20 & 0.8632 \\
\rowcolor{rowours}
\textbf{FG + QG} & 5.06 & 39.53 & 181.66 & \textbf{28.55} & \textbf{0.8688} \\
\bottomrule
\end{tabularx}
\end{table}

\Needspace{5\baselineskip}
To test whether QG improves reconstruction when given the same number of updates, we use identical feature grid configurations, two hidden layers with 64 neurons each, and uniform sampling under a common 20,000-step learning-rate schedule. This schedule also applies to the sampling comparison below. Tables~\ref{tab:ablation_comparison} and~\ref{tab:sampling_combinations} report global video PSNR and mean frame-wise MS-SSIM evaluated on unquantized FP32 reconstructions, avoiding additional distortion from quantization. Fitting time includes compilation, while decoding efficiency is measured using warmed-up forward-pass throughput.

Table~\ref{tab:ablation_comparison} shows that replacing the MLP decoder with QG improves PSNR and MS-SSIM on both sequences, with a larger PSNR gain on \textit{Jockey}. The parameter count changes only slightly because the feature grids account for most of the model parameters. This is consistent with the motivation for gating: enriching interactions among the retrieved features without substantially increasing the total parameter count. Those interactions nevertheless require extra computation, so completing the same number of updates takes longer, and decoding throughput is lower than with the MLP decoder.

\begin{figure}[!htbp]
\centering
    \includegraphics[width=\linewidth,trim=0 2bp 0 5bp,clip]{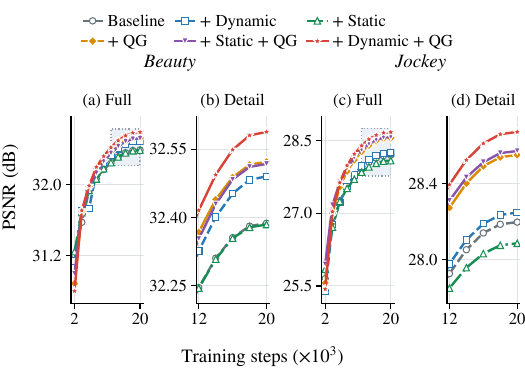}
\caption{Global video PSNR over 20,000 updates under the settings of Tables~\ref{tab:ablation_comparison} and~\ref{tab:sampling_combinations}. Panels (a--b) show the full trajectory and late-stage detail for \textit{Beauty}, and (c--d) show those for \textit{Jockey}. The baseline uses uniform sampling with an MLP decoder, whereas + QG replaces the decoder with QG while retaining the same feature grids.}
\label{fig:convergence_curves}
\end{figure}

Figure~\ref{fig:convergence_curves} shows how reconstruction quality develops during fitting. Under uniform sampling, QG reaches the MLP baseline's final PSNR with fewer updates on both sequences and continues to improve. Together, the compact decoder comparison and the full-architecture ablation support the use of QG in Q-GAS, as it improves reconstruction quality while maintaining real-time decoding capability.

\begin{table}[!htbp]
\centering
\caption{Comparison of sampling strategies and their combinations with the QG decoder. Apart from the sampling strategies, the training and evaluation settings follow those in Table~\ref{tab:ablation_comparison}. FG denotes 3D feature grid. Rows without Dynamic or Static use uniform sampling. Green shading highlights the QG configurations. The best result for each metric and sequence within each decoder group is bold-faced.}
\label{tab:sampling_combinations}
\footnotesize
\setlength{\tabcolsep}{2pt}
\renewcommand{\arraystretch}{1.05}
\begin{tabularx}{\columnwidth}{>{\raggedright\arraybackslash}Xc|cccc}
\toprule
\rowcolor{headerblue}
Method & Params (M) & \makecell{Time\\(s)$\downarrow$} & FPS$\uparrow$ & \makecell{PSNR\\(dB)$\uparrow$} & MS-SSIM$\uparrow$ \\
\midrule
\multicolumn{6}{c}{\cellcolor{gray!15}\textbf{\textit{Beauty}}} \\
\midrule
FG + MLP & 5.05 & 32.48 & 400.43 & 32.39 & 0.8833 \\
FG + Dynamic + MLP & 5.05 & 36.82 & 405.86 & \textbf{32.49} & \textbf{0.8839} \\
FG + Static + MLP & 5.05 & \textbf{32.16} & \textbf{406.62} & 32.38 & 0.8829 \\
\midrule
\rowcolor{rowours}
\textbf{FG + Dynamic + QG} & 5.06 & 42.29 & \cellcolor{bestcell}{\textbf{181.22}} & \cellcolor{bestcell}{\textbf{32.59}} & \cellcolor{bestcell}{\textbf{0.8849}} \\
\rowcolor{rowours}
\textbf{FG + Static + QG} & 5.06 & \cellcolor{bestcell}{\textbf{39.06}} & 180.51 & 32.52 & 0.8841 \\
\midrule
\multicolumn{6}{c}{\cellcolor{gray!15}\textbf{\textit{Jockey}}} \\
\midrule
FG + MLP & 5.05 & 32.91 & 407.66 & 28.20 & \textbf{0.8632} \\
FG + Dynamic + MLP & 5.05 & 36.05 & 407.19 & \textbf{28.25} & 0.8607 \\
FG + Static + MLP & 5.05 & \textbf{31.78} & \textbf{411.48} & 28.09 & 0.8562 \\
\midrule
\rowcolor{rowours}
\textbf{FG + Dynamic + QG} & 5.06 & 40.76 & 181.76 & \cellcolor{bestcell}{\textbf{28.67}} & \cellcolor{bestcell}{\textbf{0.8684}} \\
\rowcolor{rowours}
\textbf{FG + Static + QG} & 5.06 & \cellcolor{bestcell}{\textbf{39.14}} & \cellcolor{bestcell}{\textbf{186.59}} & 28.57 & 0.8663 \\
\bottomrule
\end{tabularx}
\end{table}

\Needspace{5\baselineskip}
\noindent\textbf{Effect of Sampler Selection.} Having selected the decoder, we now examine how training coordinates should be sampled. Table~\ref{tab:sampling_combinations} first compares uniform, dynamic, and static sampling with the same MLP decoder, then evaluates dynamic and static sampling with QG. This comparison isolates the effect of sampling under the same decoder architecture and then examines its interaction with QG.

With the MLP decoder, dynamic sampling yields modest PSNR gains over uniform sampling on both sequences. On \textit{Jockey}, however, its MS-SSIM is slightly lower. Static sampling remains close to uniform sampling on \textit{Beauty} but lowers both quality measures on \textit{Jockey}. The sampling criteria may help explain this contrast. Static sampling prioritizes adjacent-frame changes, which need not match the model's current reconstruction errors. Dynamic sampling instead adapts its priorities to these errors as fitting progresses.

With QG, dynamic sampling yields the highest final PSNR on both sequences and improves MS-SSIM over static sampling. Uniform-QG still has slightly higher MS-SSIM on \textit{Jockey}, so the PSNR advantage does not imply an improvement in every aspect of fidelity. Importantly, these PSNR gains come with little change in decoding throughput when the decoder architecture is unchanged. Sampling changes which coordinates contribute to each training update, while leaving the inference architecture and reconstruction operations unchanged. Maintaining the error map and refreshing the candidate pool instead add work during training, consistent with the longer fitting times under the common update budget.

\begin{figure}[!htbp]
    \centering
    % Equal image heights and shared row coordinates align both GT views with the reconstructions.
    \begin{tikzpicture}[
        image/.style={anchor=north,inner sep=0},
        panel label/.style={anchor=north,inner sep=0,font=\fontsize{8}{9.5}\selectfont,
            align=center,text width=0.215\linewidth,yshift=-2pt}]
        \node[image] (gtframe) at (0.17\linewidth,0)
            {\includegraphics[height=0.19\linewidth]{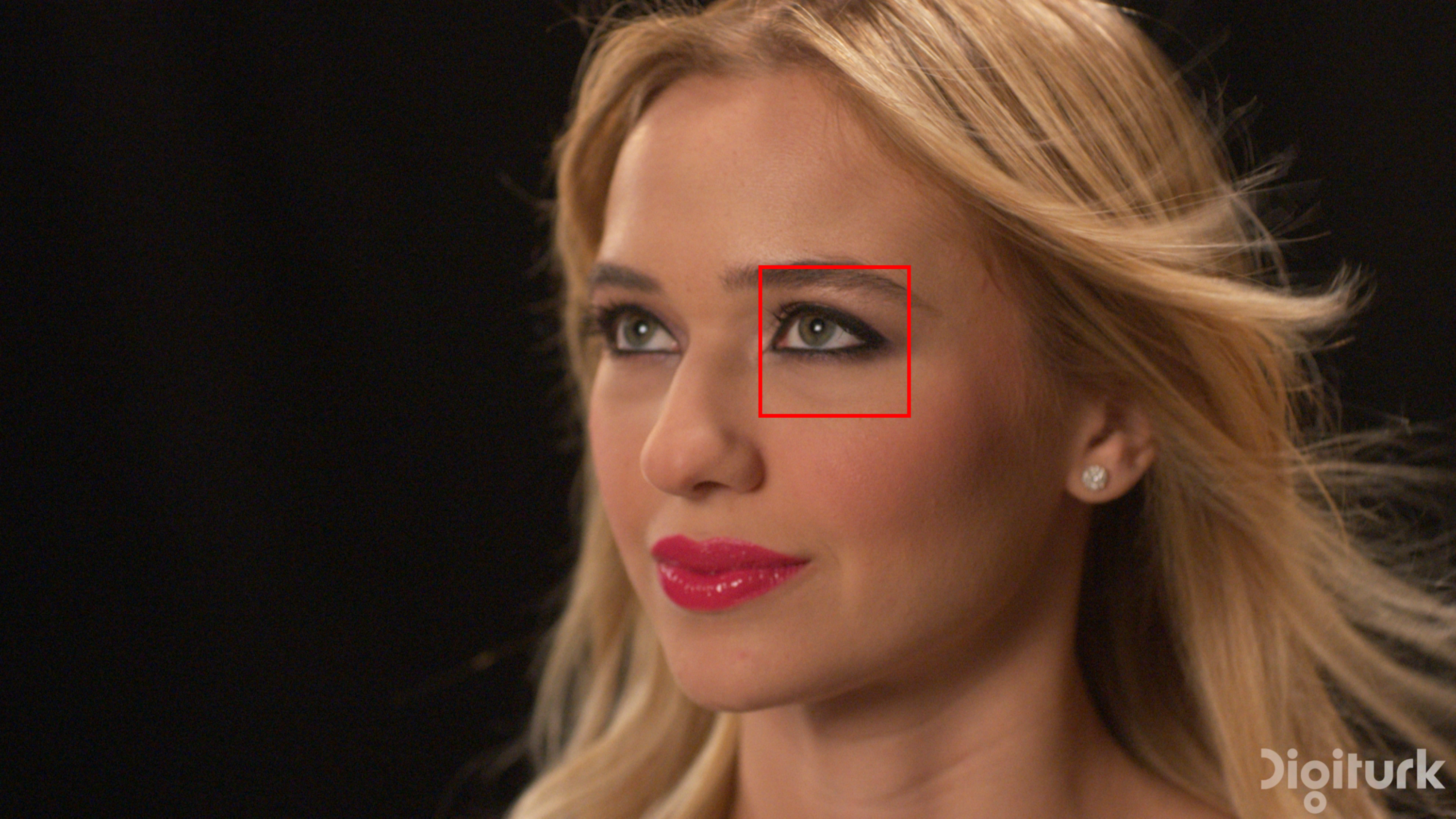}};
        \node[image] (mlp) at (0.445\linewidth,0)
            {\includegraphics[height=0.19\linewidth]{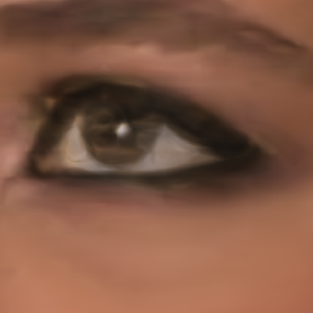}};
        \node[image] (dynamicmlp) at (0.6675\linewidth,0)
            {\includegraphics[height=0.19\linewidth]{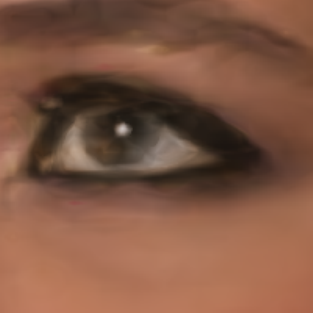}};
        \node[image] (staticmlp) at (0.89\linewidth,0)
            {\includegraphics[height=0.19\linewidth]{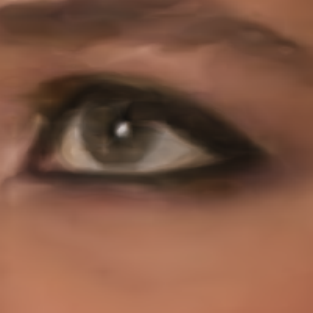}};
        \node[image,yshift=-15pt] (gteye) at (0.17\linewidth,-0.19\linewidth)
            {\includegraphics[height=0.19\linewidth]{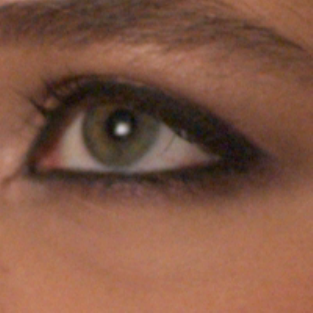}};
        \node[image,yshift=-15pt] (qg) at (0.445\linewidth,-0.19\linewidth)
            {\includegraphics[height=0.19\linewidth]{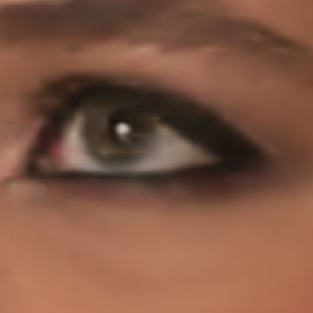}};
        \node[image,yshift=-15pt] (staticqg) at (0.6675\linewidth,-0.19\linewidth)
            {\includegraphics[height=0.19\linewidth]{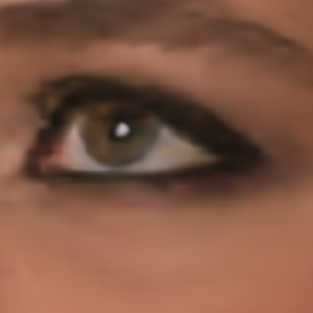}};
        \node[image,yshift=-15pt] (dynamicqg) at (0.89\linewidth,-0.19\linewidth)
            {\includegraphics[height=0.19\linewidth]{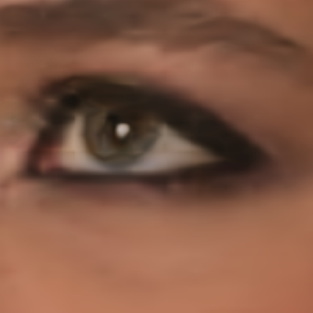}};
        % Uniform single-line labels with protected method names and equal baselines.
        \node[panel label] at (mlp.south) {\strut\mbox{(b)\,FG+MLP}};
        \node[panel label] at (dynamicmlp.south) {\strut\mbox{(c)\,FG+D+MLP}};
        \node[panel label] at (staticmlp.south) {\strut\mbox{(d)\,FG+S+MLP}};
        \node[panel label,text width=0.34\linewidth] at (gteye.south) {\strut\mbox{(a)\,Ground Truth}};
        \node[panel label] at (qg.south) {\strut\mbox{(e)\,FG+QG}};
        \node[panel label] at (staticqg.south) {\strut\mbox{(f)\,FG+S+QG}};
        \node[panel label] at (dynamicqg.south) {\strut\mbox{(g)\,\textbf{FG+D+QG}}};
        % Original red box: x=1000..1200, y=350..550 in a 1920x1080 frame.
        \begin{scope}[shift={(gtframe.south west)},
            x=0.337778\linewidth,y=0.19\linewidth]
            \draw[red,line width=0.35pt,dash pattern=on 1.5pt off 1pt]
                (0.520833,0.490741) -- (gteye.north west);
            \draw[red,line width=0.35pt,dash pattern=on 1.5pt off 1pt]
                (0.625,0.490741) -- (gteye.north east);
        \end{scope}
    \end{tikzpicture}
    \caption{Visual comparison on \textit{Beauty}. Left: the ground-truth frame and enlarged eye region linked by red dashed lines. Right: reconstructed eye regions. FG denotes 3D feature grids, and D/S denote dynamic/static sampling. Configurations without D/S use uniform sampling. These examples are from earlier runs, separate from Tables~\ref{tab:ablation_comparison} and~\ref{tab:sampling_combinations}.}
    \label{fig:visual_comparison}
\end{figure}

\begin{figure}[!htbp]

    \centering
    \includegraphics[width=\linewidth,trim=0 3bp 0 6bp,clip]{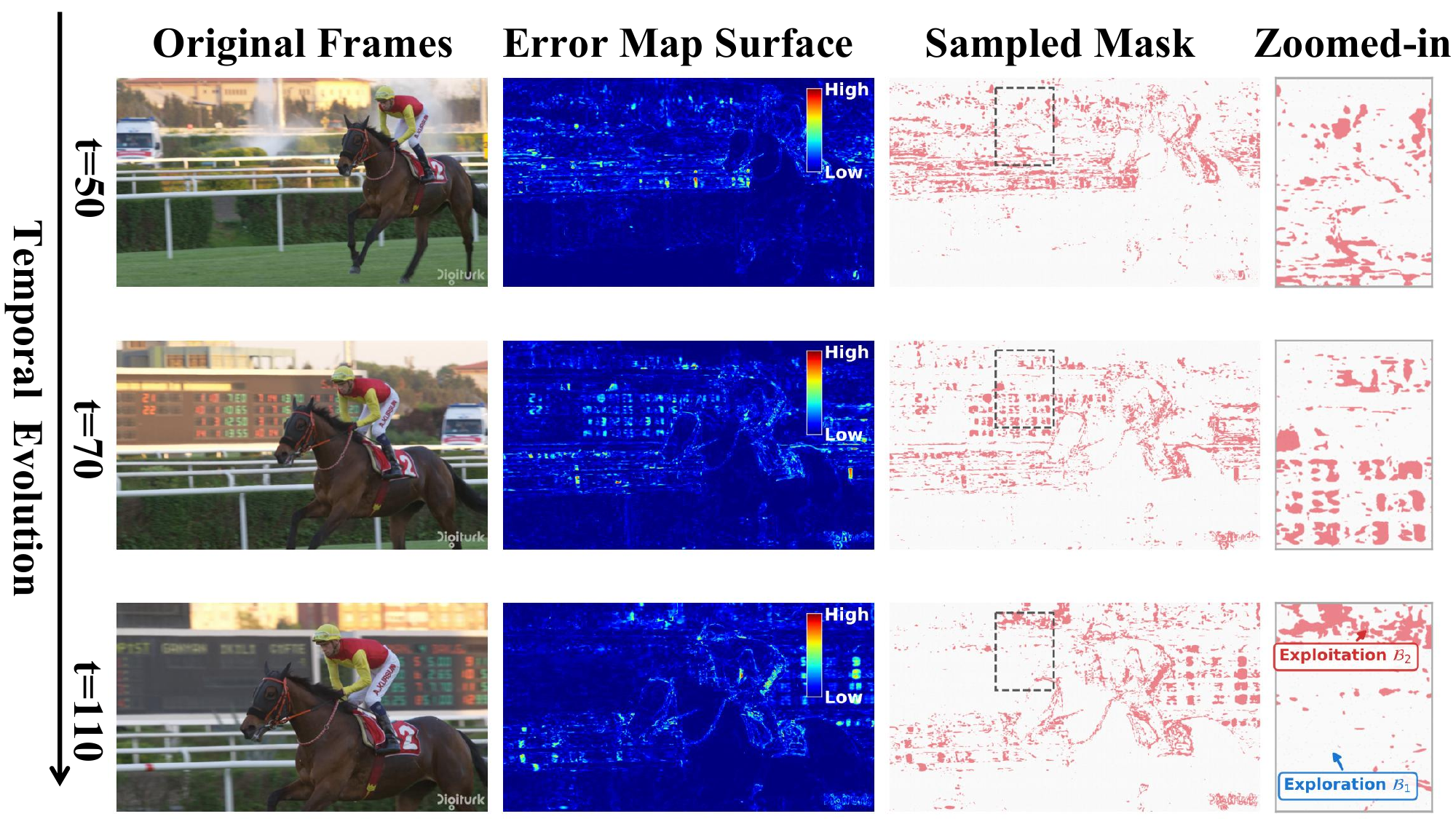}
    \caption{Dynamic error-aware sampling on \textit{Jockey}. The latest error map $\mathcal{E}$ guides the prioritized samples in $\mathcal{B}_2$ toward high-error regions, including the horse and rider, while $\mathcal{B}_1$ provides uniform samples from the full sequence.}
    \label{fig:error_surface}
\end{figure}

\Needspace{3\baselineskip}
The enlarged panels in Figure~\ref{fig:convergence_curves} show that Dynamic--QG maintains its PSNR advantage over Static--QG throughout the later updates, rather than only at the final checkpoint.  These trajectories complement the fixed-time results in Section~\ref{sec:comparative_evaluation}, where Q-GAS with dynamic sampling also retains a quality advantage within the available training budget.

\Needspace{3\baselineskip}
Figure~\ref{fig:visual_comparison} further compares the reconstructed details in an enlarged eye region. The MLP reconstruction exhibits streaks around the iris and the bright reflection. The dynamic MLP variant reduces some of these artifacts, while the QG combinations more closely match the ground truth by recovering a more coherent iris boundary and a more localized reflection.

Figure~\ref{fig:error_surface} shows how the resulting candidate pool concentrates prioritized samples in regions with large stored errors, while the uniform branch preserves sequence-wide sampling support. Together, these analyses support complementary roles for QG and dynamic sampling: the decoder enriches feature interactions, while the sampler allocates additional training samples to coordinates with larger reconstruction errors. We therefore adopt this combination as the default configuration for Q-GAS, balancing reconstruction quality, fitting time, and decoding efficiency for video compression.

\Needspace{5\baselineskip}
\section{Conclusion}

In this work, we have presented Q-GAS, a continuous video representation that couples multi-resolution 3D feature grids with a compact QG decoder. The grids store independent vertex features, while QG introduces nonlinear feature interactions with limited parameter overhead. During training, dynamic error-aware sampling prioritizes coordinates with large reconstruction errors while retaining sequence-wide uniform sampling. Experiments on UVG sequences have shown that Q-GAS fits quickly, decodes in real time, and achieves competitive reconstruction quality. In the matched-size comparison, Q-GAS completes training and quantization more than 5$\times$ faster than the evaluated Instant-NGP implementation and achieves over 20$\times$ its decoding throughput. Under the fixed training budget, Q-GAS with dynamic sampling achieves higher reconstruction quality than the evaluated NeRV-family baselines across their overlapping BPP ranges. Ablation studies further show that QG improves reconstruction quality, while dynamic sampling provides additional PSNR gains without altering the inference architecture. Future work will explore more compact and more expressive grid representations to further accelerate fitting and improve reconstruction quality.

\bibliographystyle{IEEEtran}
\bibliography{Q-GAS}

\end{document}